\documentclass[10pt,twocolumn,letterpaper]{article} %

\usepackage[pagenumbers]{cvpr} 
\usepackage{appendix}
\usepackage{graphicx}
\usepackage{amsmath}
\usepackage{amssymb}
\usepackage{comment}
\usepackage{color}
\usepackage{multirow}
\usepackage[misc]{ifsym}
\usepackage{caption}
\usepackage{array}
\usepackage{algorithm}
\usepackage{algorithmic}
\usepackage{float}
\usepackage{multirow}

\usepackage[pagebackref,breaklinks,colorlinks]{hyperref}

\usepackage[capitalize]{cleveref}
\crefname{section}{Sec.}{Secs.}
\Crefname{section}{Section}{Sections}
\Crefname{table}{Table}{Tables}
\crefname{table}{Tab.}{Tabs.}

\def\confName{CVPR}
\def\confYear{2022}

\begin{document}

\captionsetup[figure]{labelfont={bf},labelformat={default},labelsep=period,name={Fig.}}
\title{Prior-Guided One-shot Neural Architecture Search}

\author{Peijie Dong$^{1}$, Xin Niu$^{1}$, Lujun Li$^{2}$, Linzhen Xie$^{1}$, Wenbin Zou$^{3}$, Tian Ye$^{4}$, Zimian Wei$^{1}$, Hengyue Pan$^{1}$\\
$^1$ School of Computer, National University of Defense Technology, Hunan, China\\
$^2$ Chinese Academy of Sciences, Beijing, China\\
$^3$ Fujian Provincial Key Laboratory of Photonics Technology, Fujian Normal University, Fuzhou, China\\
$^4$ School of Ocean Information Engineering, Jimei University, Xiamen, China\\
\small  $\left\{ dongpeijienudt, niuxin,weizimian16, hengyuepan \right\}$ \small @nudt.edu.cn \\
\small lilujunai@gmail.com, 18810698745@163.com, alexzou14@foxmail.com, 201921114031@jmu.edu.cn}


\maketitle

\begin{abstract}
Neural architecture search methods seek optimal candidates with efficiency weight-sharing supernet training. However, recent studies indicate poor ranking consistency about the performance between stand-alone architectures and shared-weight networks. In this paper, we present Prior-Guided One-shot NAS (PGONAS) to strengthen the ranking correlation of supernets. Specifically, we first explore the effect of activation functions and propose a balanced sampling strategy based on the Sandwich Rule to alleviate weight coupling in the supernet. Then, FLOPs and Zen-Score are adopted to guide the training of supernet with ranking correlation loss. Our PGONAS ranks the 3rd place in the supernet Track of CVPR2022 Second lightweight NAS challenge.  The code is available at https://github.com/pprp/CVPR2022-NAS-competition-Track1-3th-solution.


 \end{abstract}


\section{Introduction}


There are tremendous advances in deep learning, including automated machine learning. Neural Architecture Search (NAS) is a branch of automated machine learning (AutoML) that has sparked increased interest due to its remarkable progress in a variety of computer vision tasks\cite{zhang2021semi,liu2020block,chen2020automated,ding2021bnas,liu2019auto,jiang2021learning}. By balancing performance and resource constraints, it aims to reduce the cost of human efforts in manually designing network architectures and discover promising models automatically. Many early NAS works are based on Reinforcement Learning and evolutionary algorithms. However, these techniques~\cite{ye2020distributed,zhong2020blockqnn} require sampling and evaluating numerous network architectures from the search space, which can take hundreds of days with thousands of GPUs.

\begin{figure}
    \centering
    \includegraphics[width=0.99\linewidth]{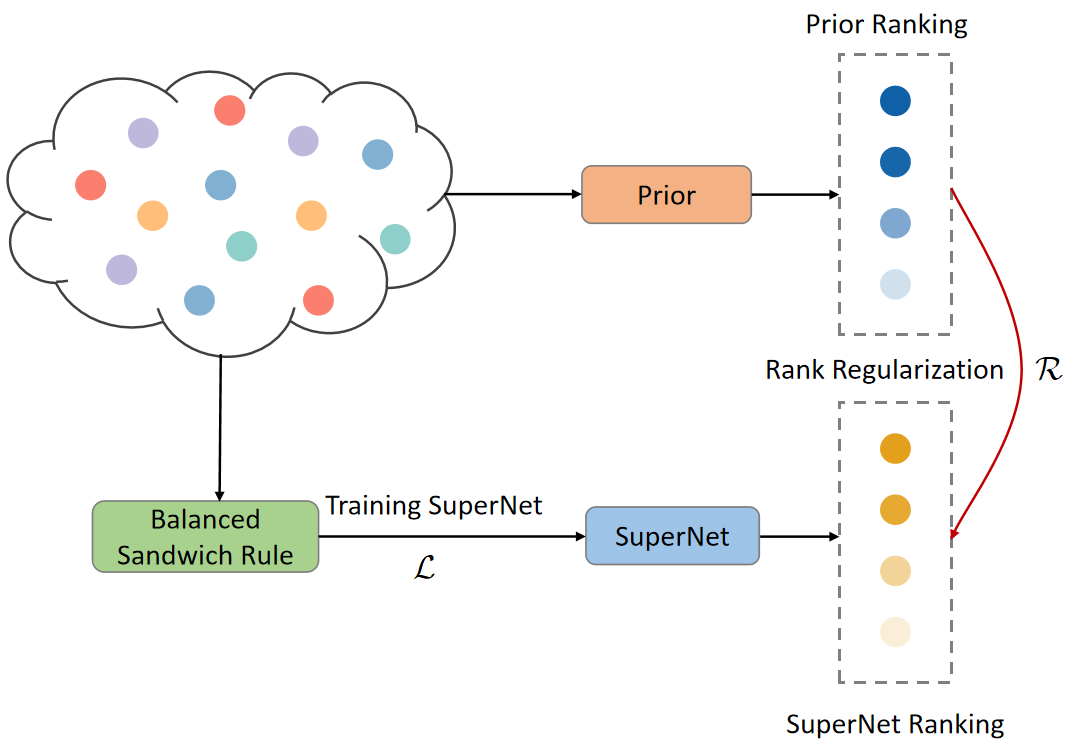}
    \caption{Overview of the Prior-Guided One-shot NAS (PGONAS). The Balanced Sandwich Rule means that the proposed balanced sampling strategy is based on the Sandwich Rule\cite{Yu2020BigNASSU}. During the training phase, priors are employed to guide the supernet training with rank loss.} \label{rankloss}
\end{figure}

To alleviate this challenge, ENAS~\cite{pham2018efficient} is the first to present a weight-sharing mechanism for efficient NAS to optimize computation. It defines the supernet in which all models share a single copy of weights, rather than training models from scratch. DARTS \cite{liu2018darts} and DARTS-like methods~\cite{wu2019fbnet, xu2021partially, cai2018proxylessnas} are another popular type of weight-sharing approach. They optimize the continuous architecture parameters and network weights during supernet training. In contrast, One-shot NAS\cite{guo2019single, zhang2020one, dong2019one} is a new paradigm that separates architecture search from supernet training. One-shot NAS uses evolutionary algorithms to sample numerous architectures in the training stage of supernet. Then, these candidates are then evaluated by inheriting weights from the well-trained supernet in search stage.  And finally, we train the best performers. It requires less memory and is more efficient because only a portion of the candidates are activated and optimized.

Despite the fact that One-shot NAS significantly improves search efficiency, it is prone to poor performance estimation during the evaluation process. It typically searches for a large and complex network space with billions of options to find the best ones, in which the supernet parameters are tightly coupled. When child models are sampled and trained in different iterations, they interfere with each other, and their accuracy in the supernet is unavoidably averaged, blurring the lines between strong and weak architectures. As a result, a well-trained supernet finds it difficult to obtain a stable and accurate performance ranking of candidate models. 

In this paper, we propose Prior-Guided One-shot NAS (PGONAS), an effective strategy to address the weak correlation problem of weight-sharing supernet. The PGONAS improvement for One-shot NAS can be highlighted in three aspects: architecture enhancement, sampling strategies, and prior-based regularization. First, we utilize PReLU to replace the second activation of the block to effectively enhance the predictability of the supernet in the channel dimension. Then, unlike the naive sampling strategy, we use the Sandwich Rule (maximum, middle, and minimum) sampling strategy and in place distillation. These two strategies achieve 81.56 absolute correlation coefficient, while the baseline is 72.49. Encouraged by the recent train-free NAS, we also evaluate some train-free metrics and are surprised with its relatively advanced consistency (78.43 for FLOPs and 81.29 for Zen-Score). Therefore, we introduce these train-free metrics to guide the training of the supernet as a priori. Specifically, we add a consistent regularization loss for arbitrary pairs of candidates about its train-free metrics and the training losses on the supernet. To further improve, we carefully tune the weights of the loss and architectural distances. Finally, with the above advanced supernet training techniques, we obtain the consistency of 83.20 and ranked 3rd in the supernet track of the CVPR 2022 NAS Challenge.

\section{Related Work}
\label{sec:related_work}
In this section, we briefly summarize the investigative techniques behind our approach, including One-shot NAS and prior metrics.

\noindent\textbf{One-shot NAS.}
In a wide variety of computer vision tasks~ \cite{hu2018manifold, fu2020scene, zhao2019object}, manually designed neural networks had great success. Artificial architectures, on the other hand, are commonly believed to be sub-optimal. Both academia and industry have recently become more interested in neural architecture search (NAS). The early NAS work trains the child networks individually by RNN and reinforcement learning, but consumes a large computer resource. Recent One-shot NAS utilize weight-sharing super-network for efficient NAS to reduce computation costs. Different child models in these algorithms share the same weights. They initially train an over-parameterized supernet utilizing different sample strategies. After that, search a discrete search space with numerous candidates. Sampling strategies are important in the training stage because they affect how to train an accurate and stable super-network for performance estimation. \cite{bender2018understanding}, for example, trains the over-parameterized network while dropping out operators with increasing probability, allowing their weights to co-adapt. SPOS~\cite{guo2019single} proposes the uniform sampling method for supernet training based on this. Only one path is activated for each optimization step, and regular gradient-based optimizers are used to optimize it. FairNAS~\cite{chu2019fairnas} strengthens the One-shot method by adhering to strict fairness for both super-network sampling and training. AutoSlim\cite{Yu2019AutoSlimTO} improves correlation by optimizing the maximum \& minimum and intermediate paths by in-place distillation. OFA proposes a one-stage supernet training strategy through progressive shrinkage. As an alternative technology paradigm, the  gradient-based methodologies~\cite{liu2018darts,wu2019fbnet}  initiate architecture parameters with each operator, including using back-propagation to jointly optimize them and network weights. Finally, magnitudes of architecture parameters are used to select the best model. In contrast, we propose a new One-shot NAS method based on Autoslim\cite{Yu2019AutoSlimTO} \& priori-guided regularization, which is not present in previous work.

\noindent\textbf{Training-free Metric.}
In recent years, many researchers began to realize that some training-free metrics can be used to measure the capacity of neural networks. These methods are generally based on Gaussian initialization of the model and the ability to characterize random features. Specifically, TE-NAS estimates expressiveness by directly computing the number of active regions RN in randomly sampled images. The number of active regions is calculated directly on the images. NASWOT calculates the architecture score based on the kernel matrix of binary activation patterns between small batches of samples. Zen-score takes into account not only the distribution of linear regions, but also the complexity, resulting in a more accurate estimate of the expressiveness of the network. In addition, the number of parameters and the computational volume of the model are also indicators that can initially represent the capacity of the model. In many scenarios, the larger model performs better. In this work, we select FLOPs and Zen-score to guide the training of the supernet using ranking losses.

\section{Prior-guided One-shot NAS}
In this section, we present our architecture enhancement $\&$ sampling strategies for weight-sharing supernet, and ranking regularization based on a priori metrics, respectively.
\subsection{Weight-sharing supernet Enhancement}

Weight sharing in the supernet can lead to gradient conflicts, making the supernet difficult to converge and affecting the ranking consistency of the supernet. Therefore, in this paper, we explore the effects of activation functions, sampling strategies, and regularization on the ranking consistency of the supernet.

\begin{figure}
    \centering
    \includegraphics[width=4cm]{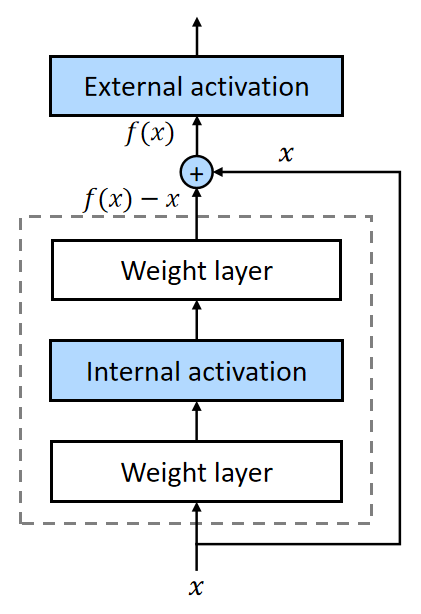}
    \caption{Illustration of the location of activations in residual blocks.} \label{activations}
\end{figure}

\noindent\textbf{Activation Functions.} Inspired by \cite{Yang2021ImprovingRC}, we explore the effects of activation functions in detailed, including the location and type of activation functions. ReLU is a nonlinear activation function that allows complex patterns in the data to be learned. However, ReLU cannot learn examples for which their activation is zero. If the input is less than 0, then it outputs zero, and the neural network cannot continue the back propagation algorithm, which is known as the dying ReLU problem. We observe that replacing ReLU with smoother activation functions helps alleviate the ranking disorder problem. Take Mish for example, it provides a strong regularization effect and helps make gradients smoother, which makes it easier to optimize function contour. On the other hand, smoother activation functions have wider minima, which improve generalization compared to ReLU. In addition, the location of the activation functions also matters. As shown in Figure \ref{activations}, there are two activation functions in the residual block. Define the activation function between the two weight layer as internal activation, and the activation function after shortcut as external activation. We found that internal activation plays an important role in supernet training. If we replace the internal ReLU activation with smoother activation functions such as PReLU, the supernet would suffer from gradient explosion and is difficult to converge. Also, the supernet is not sensitive to converting the external activation functions.

\begin{figure}
    \centering
    \includegraphics[width=0.99\linewidth]{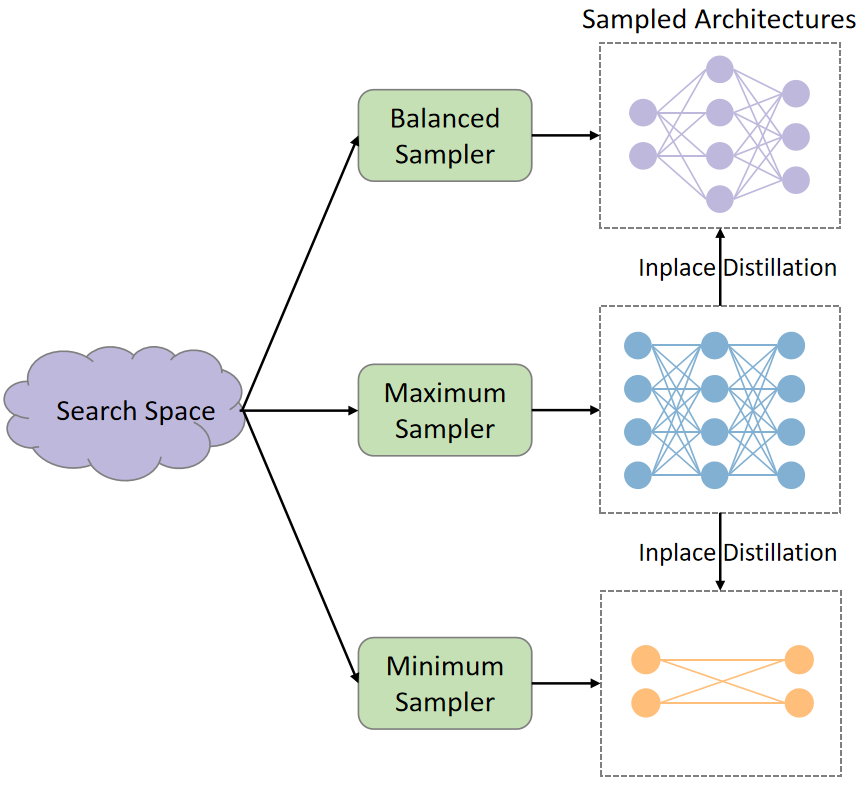}
    \caption{Illustration of the Sandwich Rule with Balanced Sampler.}\label{Sandwich}
\end{figure}

\noindent\textbf{Sampling Strategies.} Two-stage NAS requires sampling from the search space during training. Thus, sampling strategies would directly impact supernet optimization. A suitable sampling strategy can alleviate the interference between sub-networks in the supernet. There are many sampling strategies in One-shot NAS, such as the uniform sampling strategy in SPOS \cite{guo2019single}, the progressive shrinking strategy in Once for all \cite{cai2019once}, and the Sandwich Rule in Autoslim \cite{Yu2019AutoSlimTO} and BigNAS \cite{Yu2020BigNASSU}. Among them, we found that the Sandwich Rule achieves a better rank correlation in our search space. we proposed a noval sampling strategy based on the Sandwich Rule. Specifically, we replace the original random sampler with a balanced sampler, which is motivated by BNAO \cite{Luo2019BalancedON}.  Training of architectures of different sizes is imbalanced, which causes the evaluated performances of the architecture to be less predictable of their stand-alone accuracy. BNAO proposed sampling based on the model size of sub-networks. As shown in Figure \ref{Sandwich}, there are three samplers in PGONAS: 

\begin{enumerate}
   \item Maximum Sampler: sample the subnet with the largest width and the largest depth. 
   \item Minimum Sampler: sample the subnet with the smallest width and smallest depth.
   \item Balanced Sampler: sample N subnets with random width and random depth and sample one of them based on $p_{i} = \frac{FLOPs_{i}}{\sum_j{FLOPs_{j}}}$.
\end{enumerate}

In-place distillation is also adopted in PGONAS. The largest model is training with true label, the smallest model and the middle size model are trained with in-place distillation.

\subsection{Prior-guided Ranking Regularization}

\noindent\textbf{Zero-shot Prior}
We adopt FLOPs and Zen-Score as prior, respectively, to guide the training procedure. With the efficient zero-shot NAS\cite{Lin2021ZenNASAZ, Chen2021NeuralAS, Mellor2021NeuralAS}, the search cost has been greatly reduced. 
Among them, we finally choose the Zen-Score\cite{Lin2021ZenNASAZ} as prior because of its stable and scale-insensitive estimation ability. Zen-Score is a zero-shot predictor for ranking architectures, which can stably correlate with the model accuracy without training the network parameters. Zen-Score can measure network expressivity by averaging the Gaussian complexity of the linear function in each linear region. In practice, the feature map before the global average pool layer (pre-GAP) is adopted to avoid information loss. 

\noindent\textbf{Computation Prior}
Motivated by NATSBench\cite{Dong2021NATSBenchBN}, we found that the FLOPs of each subnet have a good correlation with the accuracy of the standalone model. Using FLOPs as prior to measure rank consistency is intuitive. Models with high FLOPs usually have larger capacity and are likely to achieve better performance.

\noindent\textbf{Pairwise Rank Loss} We first revisit the ranking consistency problem in One-shot NAS. Let $\Omega$ be the search space, defined as $N$ candidate architectures $a_i, i \in [1, N]$. Pairwise rank loss is used to constrain the optimization process of supernet as a regular term. Assuming that the architecture $a_i$ is better than the performance of the architecture $a_j$, then with the above ranking constraints, the following relationship is theoretically satisfied.

\begin{equation}
    \mathcal{L}_{\mathcal{A}}\left(x, \theta_{a_{i}}^{*}\right) \leq \mathcal{L}_{\mathcal{A}}\left(x, \theta_{a_{j}}^{*}\right) 
    \Rightarrow 
    \mathcal{L}\left(x, \theta_{a_{i}}^{s}\right) \leq \mathcal{L}\left(x, \theta_{a_{j}}^{s}\right)
\end{equation}

where $\mathcal{L}_\mathcal{A}$ denotes the real accuracy corresponding to stand alone training, and $\mathcal{L}$ denotes the corresponding loss function ranking in the supernet. Then the loss functions computed by architecture $a_i$ and architecture $a_j$ are as follows:

\begin{equation}
    \mathcal{R}_{(i,j)}\left(\theta^{s}\right)=\max \left(0, \mathcal{L}\left(x, \theta_{a_{i}}^{s}\right)-\mathcal{L}\left(x, \theta_{a_{j}}^{s}\right)\right)
\end{equation}

We introduce the balanced sampling strategy with Sandwich Rule and pairwise rank loss described by Algorithm \ref{pseudo}.

\begin{algorithm}
	\renewcommand{\algorithmicrequire}{\textbf{Require:}}
	\caption{Training supernet $S$ with pairwise rank loss.}
	\label{alg1}
	\begin{algorithmic}[1]
	    \STATE Require: Define $depth$ range and $width$ range.
		\STATE Require: Define $n$ as number of sampled subnets in each iteration, $m$ as number of sampling pairs in computing rank loss, $k$ as prior in guiding rank loss.
		
        initialize supernet parameters $\theta^s$ with pretrained weights.
        \WHILE{$step$ t $<$ T}
    		\STATE Get next mini-batch of data $x$ and label $y$.
    		\STATE Execute max-network, $\hat{y}=S_{max}(x).$
    		\STATE Compute loss, $loss=criterion(\hat{y}, y).$
    		\STATE Accumulate graidents, $loss.backward().$
    		\STATE Stop graidents of $\hat{y}$ as label, $\hat{y}=\hat{y}.detach().$
    		\STATE Execute min-network, $\bar{y}=S_{min}(x).$
    		\STATE Compute loss, $loss=criterion(\hat{y}, \bar{y}).$
    		\STATE Accumulate gradients, $loss.backward().$
    		
    		\WHILE{$step$ $i$ $<$ $(n-2)$}
    		    \STATE Get n networks $a_i(i=1,..,N)$ and their FLOPs $FLOPs_i(i=1,...,N)$.
    		    \STATE Sample network based on probability $p_i=\frac{FLOPs_{i}}{\sum_j{FLOPs_{j}}}$ 
    		    \STATE Execute the sampled network, $\tilde{y}=S_{rand}(x).$
    		    \STATE Compute loss, $loss=criterion(\hat{y}, \tilde{y}).$
    		    \STATE Accumulate gradients, $loss.backward().$
    	    \ENDWHILE 
    	    
    	    \WHILE{$step$ $i < m$}
    	        \STATE Sample random network $a$ and $b$.
    	        \STATE Execute random network, $y_a=S_a(x), y_b=S_b(x).$
    	        \STATE Compute prior, $k_a=(S_a), k_b=K(S_b).$
    	        \STATE Compute loss, $loss_a=criterion(y_a, y), loss_b=criterion(y_b,y).$
    	        \STATE Compute rank loss, $loss_r=\lambda \mathcal{R}_{(a,b)}=\lambda \mathcal{L}_{r}(k_a, k_b, loss_a, loss_b).$
    	        \STATE Accumulate gradients, $loss_r.backward().$
    	    \ENDWHILE 
    		
    		\STATE Update weights, $optimizer.step().$
    		\STATE Clear graidents, $optimizer.zero\_grad().$
    		
		\ENDWHILE 
	\end{algorithmic}  \label{pseudo}
\end{algorithm}

\section{Experiments}


\noindent\textbf{Search Space.} The Search Space is builded based on the ResNet48 backbone. Both the depth and the expansion ratios are searchable. There are 4 stages in the backbone. The basic channel configuration of 4 stages is 64, 128, 256 and 512. The candidate block number of the 1st, 2nd and 4th block is ranging from 2 to 5 and the candidate block number of the 3rd block is ranging from 2 to 8.  The candidate expansion ratio is, $[1.0, 0.95, 0.9, 0.85, 0.8, 0.75, 0.7]$ and the channel number should be divided by 8. There are around $5.06 \times 10^{19}$ sub-networks in total.

\noindent\textbf{Dataset.} ImageNet-mini has 34,745 training images and 3,923 validation images in 1000 classes with resolution $224\times224$. We use the official training/validation split provided by Kaggle in our experiments. 

\noindent\textbf{Training Settings.} For all experiments, we use the SGD optimizer with momentum 0.9; weight decay 0; initial learning rate 0.001 with batch size 256. 
We use cosing learning rate decay with warmup for 5 epochs. We first pre-train the supernet for 90 epochs in ImageNet-1k, and then training supernet with the Sandwich Rule for 70 epochs. 
Following the investigation in NASVIT\cite{gong2021nasvit}, stronger regularization(e.g., large weight decay, dropout, dropPath) and stronger data augmentation schemes(e.g., CutMix, Mixup, Randaugment) are not utilized in our experiments. 

\noindent\textbf{Validation Metrics} Following the setting of CVPR2022 challenge, the correlation of the rank consistency in One-shot NAS is quantified with the Pearson correlation coefficient. In these metrics, the rank correlation is bounded between -1 (disagreement) and 1 (agreement), where for 0 there is no significant correlation.  

\subsection{Results of Prior-Guided NAS}

We evaluate the effectiveness of smoother activation functions, balanced sampling strategy and rank loss with different schedulers. 

\noindent\textbf{Results of Different Activations} The experimental results are shown in Table \ref{activations_table}, where the original ReLU activation function is replaced with multiple smoother activation functions. From the table, we observe that all of the activation functions achieve better rank consistency, which prove that information loss caused by ReLU would reduce the ranking correlation of sub-networks in supernet.

\begin{table}[H]
    \centering
    \begin{tabular}{cc}
    \hline
    Activation & Pearson Coeff. \\ 
    \hline
    ReLU       & 78.20          \\
    SELU       & 79.89          \\ 
    PReLU      & 80.65          \\
    Swish      & 80.60          \\
    Mish       & \textbf{81.56}          \\
    \hline
    \end{tabular}
    \caption{Influence of different types of activation functions.}\label{activations_table}
\end{table}

\noindent\textbf{Results of Different Sample Strategy}
The sampling strategy plays an important role in optimizing the supernet. 
The comparison results with the state-of-the-art sampling strategies are shown in Table \ref{sampling}. It is surprising that the progressive shrinking strategy in Once for all\cite{cai2019once} obtain unsatisfactory results, which may indicate that although the One-stage NAS can converge well, the ranking consistency is not guaranteed. Instead, our balanced sampling strategy based on Sandwich Rule achieve good results. Specifically, we take the number of random sampled subnets as 2 and get two pairs for rank loss.

\begin{table}[H]
    \centering
    \begin{tabular}{lc}
    \hline
    Sample Strategy       & Pearson Coeff. \\ \hline
    Uniform Sampling \cite{guo2019single}      & 72.49          \\
    Progressive Shrinking \cite{cai2019once} & 69.03          \\
    Sandwich Rule \cite{Yu2020BigNASSU}        & 78.20          \\ 
    Balanced Sampling     & \textbf{81.63}          \\
    \hline
    \end{tabular}
    \caption{Influence of different sample strategies.}\label{sampling}
\end{table}

\noindent\textbf{Results of Loss coefficient $\lambda$ and schedulers.}
In Table \ref{lambda}, we compare four different coefficient schedulers: constant regularization throughout training, warm-up scheduler that gradually increases regularization, cosine scheduler that increases and then decreases, and multistage scheduler which consists of zero stage, warm-up stage, constant stage and decreasing stage.  For the warm-up scheduler, we gradually increase the loss from 0 to $\lambda_{max}=2$ linearly in the first 20 epochs to avoid an abrupt change of the loss. In the initial phase of the training supernet, the coefficient should gradually increase to avoid an abrupt change of the loss. 
As shown in Table \ref{lambda}, the warm-up scheduler gives the best results. We used this strategy in our experiments.

\begin{figure}[H]
    \centering
    \includegraphics[width=8cm]{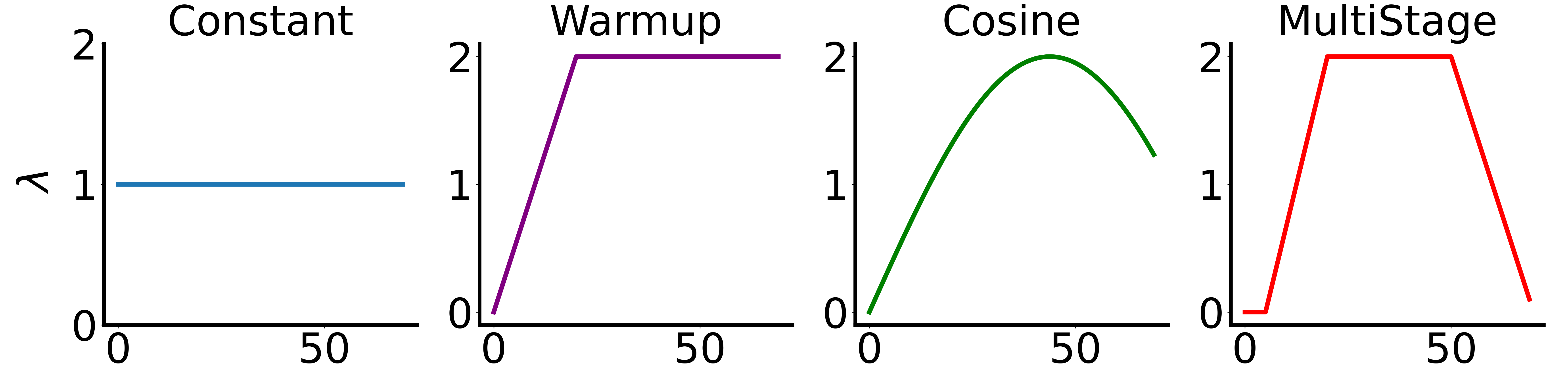}
    \caption{Different schedules of the loss coefficient $\lambda$.}
\end{figure}

\begin{table}[H]
    \centering
    \begin{tabular}{clc}
    \hline
    \multicolumn{1}{l}{Type}   & Schedule   & Pearson Coeff. \\ \hline
    \multirow{4}{*}{Rank Loss} & Constant   & 81.97          \\
                               & Warmup     & 83.20           \\
                               & Cosine     & 81.31          \\
                               & MultiStage & 81.01          \\ \hline
    \end{tabular}
    \caption{Influence of the loss coefficient $\lambda$. Different schedules (c.f. top plots) to modify the regularization throughout training.}\label{lambda}
\end{table}

\noindent\textbf{Results of different prior for guiding}
We investigate the impact of different priors in Table \ref{guiding_prior}. The Params, FLOPs and Zen-Score of sub-networks are adopted as zero-cost proxy and positively correlate with the model accuracy. Among them, FLOPs and Zen-Score achieve competitive results and are used as prior in rank loss. Performance consistently improves with the help of rank loss, which highlights the importance of prior. 

\begin{table}[H]
\centering 
\begin{tabular}{llc}
\hline
Type                          & Prior           & Pearson Coeff. \\ \hline
\multirow{3}{*}{w/o Rank Loss} & Params          & 67.23          \\
                              & FLOPs           & 78.43          \\
                              & Zen-Score        & 81.29          \\ \hline
\multirow{3}{*}{w/ Rank Loss} & w/o Prior       & 80.65          \\   
                              & FLOPs Guided    & \textbf{83.20}          \\
                              & Zen-Score Guided & 83.00          \\ \hline
\end{tabular}
\caption{Comparison of different types of prior.}\label{guiding_prior}
\end{table}

\begin{figure}[H]
\centering
\subfloat{
\begin{minipage}[t]{0.28\linewidth}
\centering
\includegraphics[width=1.0in]{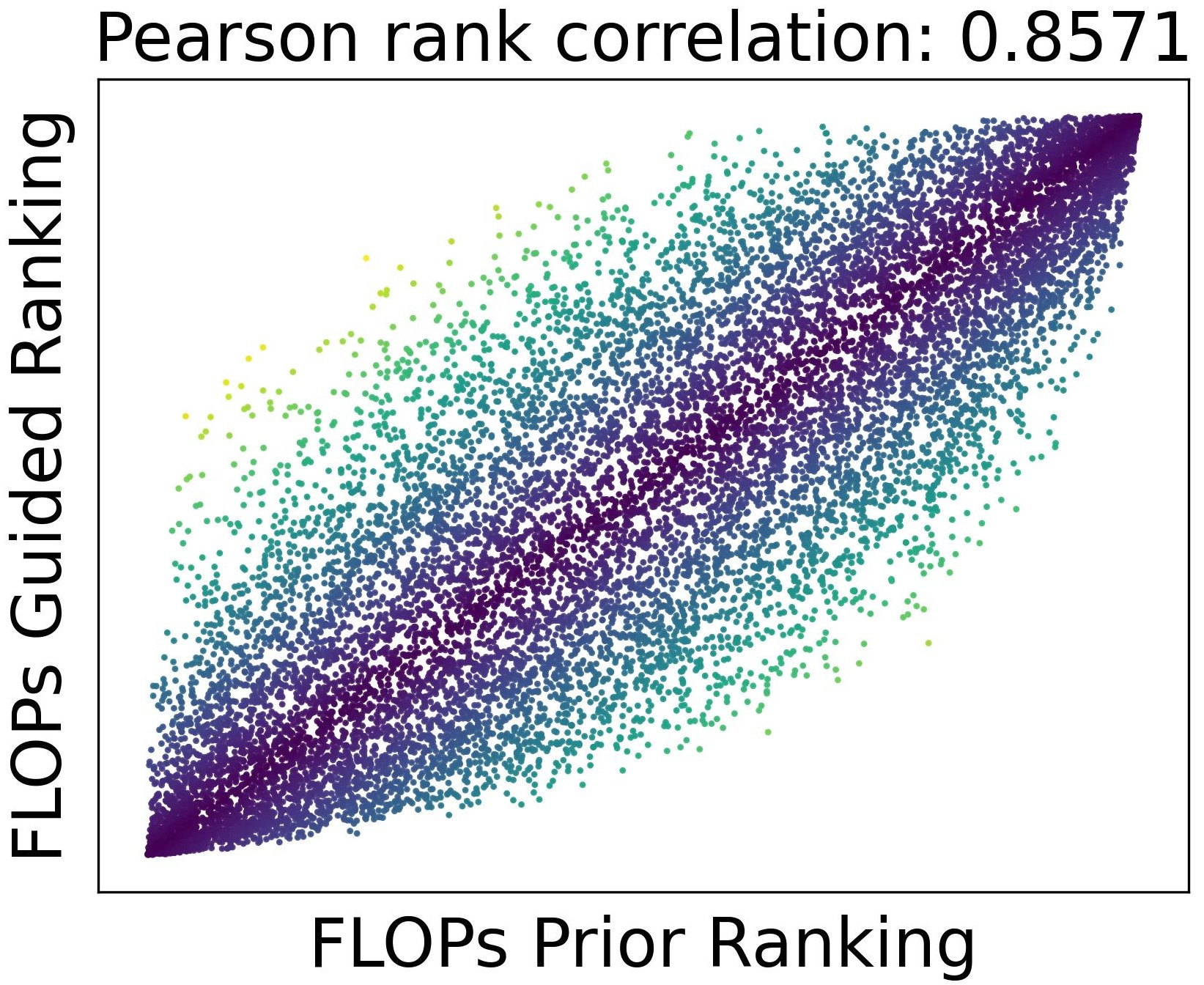}
\end{minipage}
}
\subfloat{
\begin{minipage}[t]{0.28\linewidth}
\centering
\includegraphics[width=1.0in]{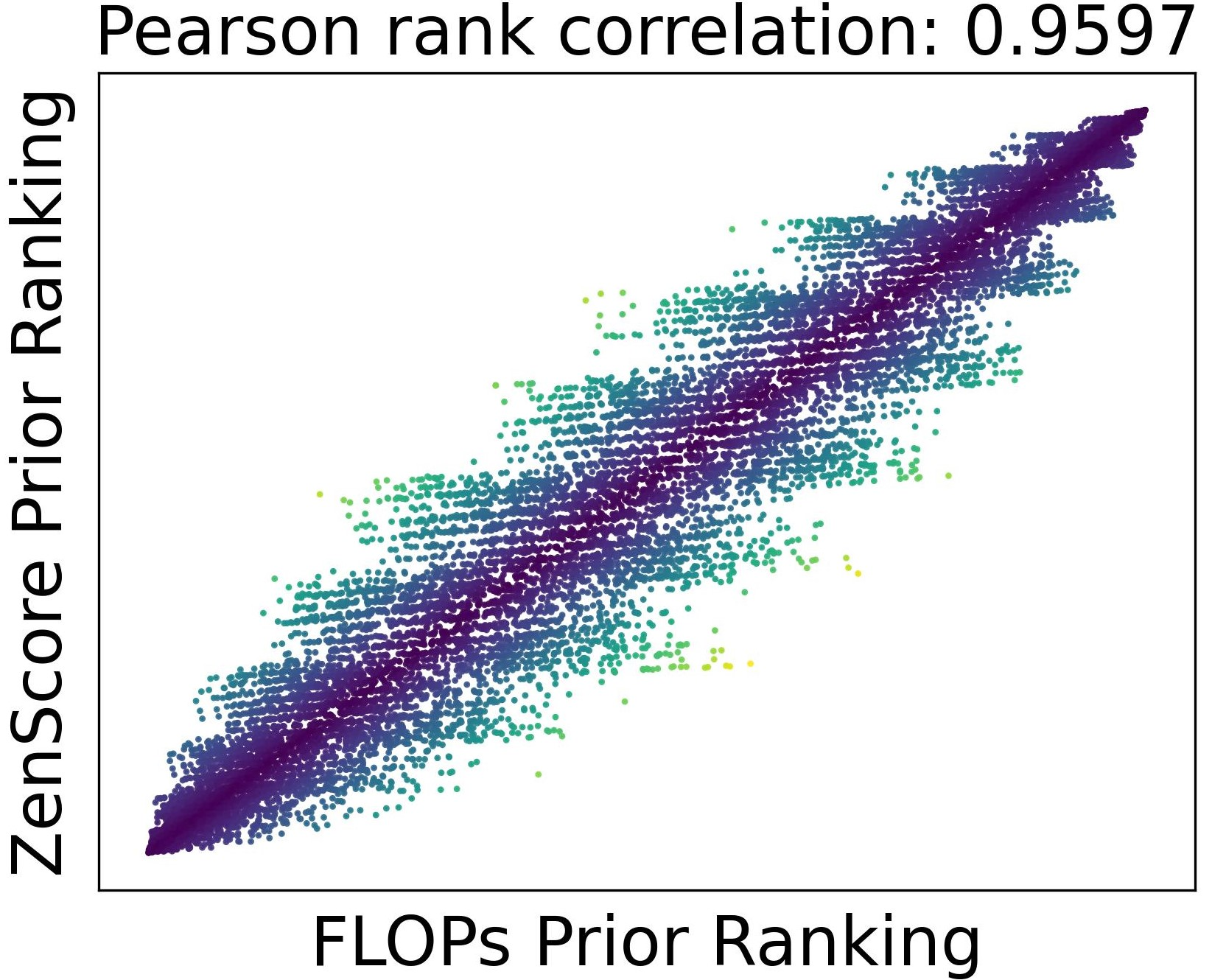}
\end{minipage}
}
\subfloat{
\begin{minipage}[t]{0.28\linewidth}
\centering
\includegraphics[width=1.0in]{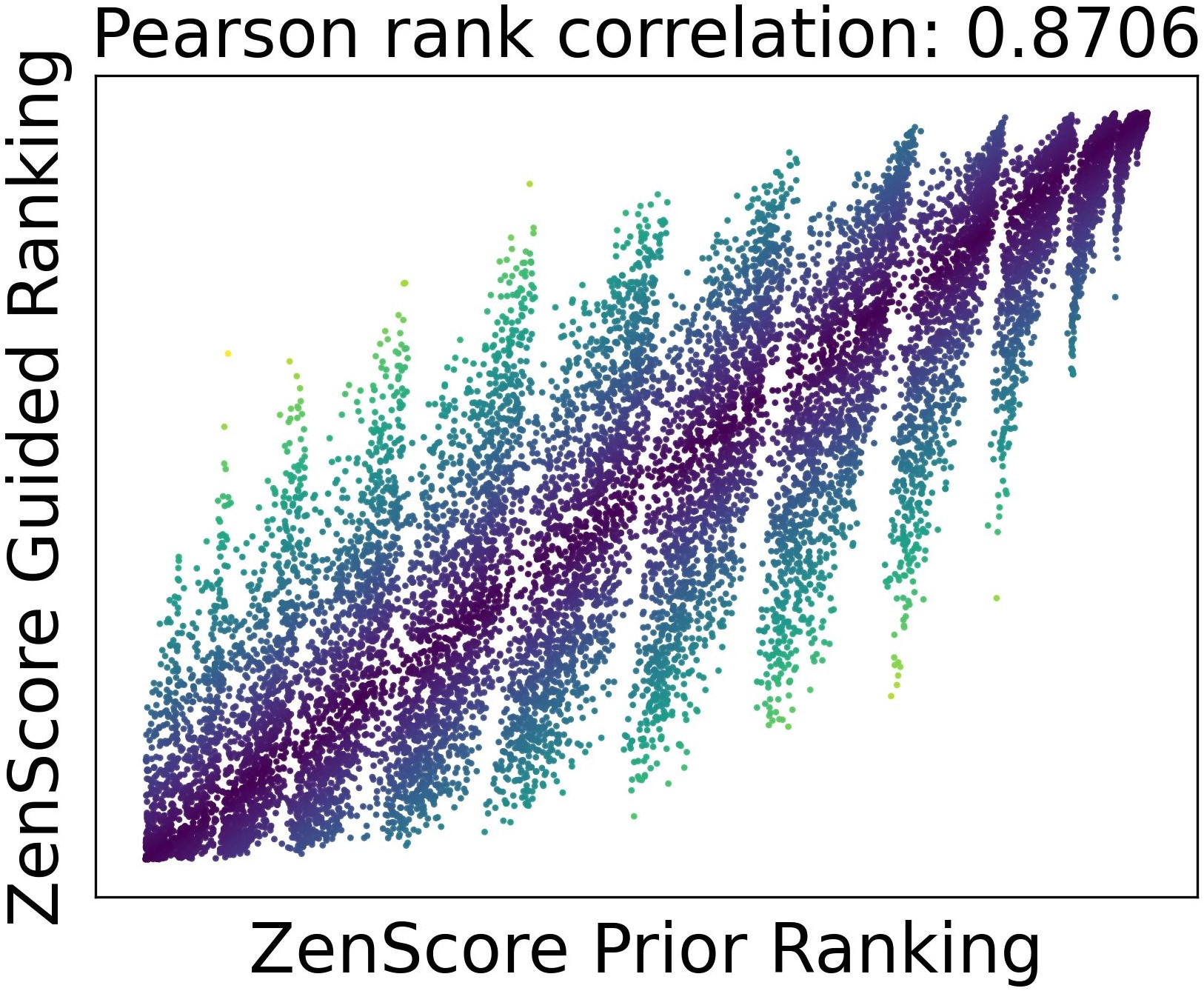}
\end{minipage}
}
\centering
\caption{Correlation between different types of priors.}
\label{correlation}
\end{figure}

\noindent\textbf{Visualization of ranking correlation}
Table \ref{guiding_prior} illustrates that the ranking correlation of Zen-Score is better than FLOPs without the guide of rank loss. However, FLOPs guided rank loss achieves a better ranking correlation than Zen-Score guided rank loss. Figure \ref{correlation} shows the correlation between different types of priors. Although the correlation between Zen-Score and FLOPs is as high as 95.97, they have different tendencies. The FLOPs proxy is continuous while the Zen-Score proxy concentrate on certain values that results in a ribbon pattern.

\section{Conclusion}

In this paper, we propose Prior-Guided One-shot NAS (PGONAS) to improve the ranking correlation of supernet. First, we explore the effect of the location and type of smoother activation functions on the ranking correlation. Second, we propose a balanced sampling strategy based on the Sandwich Rule to alleviate weight coupling in the supernet. Finally, FLOPs and Zen-Score are adopted as before to guide the training of supernet with rank loss, which consistently improves the ranking correlation. Detailed experiments demonstrate that the above methods can improve the ranking correlation of sub-networks in weights inheriting and stand-alone training. In the future, we will further explore traing-free metrics and training strategies for supernets. We hope that our approach will attract the attention of the research community for NAS and new insight.

{\small
\bibliographystyle{ieee_fullname}
\bibliography{egbib}
}

\newpage

\begin{appendices}

\end{appendices}

\end{document}